\documentclass[sigconf]{acmart}

\AtBeginDocument{%
  }

\setcopyright{acmlicensed}
\copyrightyear{2025}
\acmYear{2025}
\setcopyright{cc}
\setcctype{by}
\acmConference[UIST '25]{The 38th Annual ACM Symposium on User Interface Software and Technology}{September 28-October 1, 2025}{Busan, Republic of Korea}
\acmBooktitle{The 38th Annual ACM Symposium on User Interface Software and Technology (UIST '25), September 28-October 1, 2025, Busan, Republic of Korea}\acmDOI{10.1145/3746059.3747731}
\acmISBN{979-8-4007-2037-6/2025/09}

\usepackage{booktabs}
\usepackage{multirow}
\usepackage{bm}

\begin{document}

\title{BaroPoser: Real-time Human Motion Tracking from IMUs and Barometers in Everyday Devices}

\author{Libo Zhang}
\affiliation{%
  \institution{School of Software and BNRist, Tsinghua University}
  \country{China}
}

\author{Xinyu Yi}
\affiliation{%
  \institution{School of Software and BNRist, Tsinghua University}
  \country{China}
}

\author{Feng Xu}
\affiliation{%
  \institution{School of Software and BNRist, Tsinghua University}
  \country{China}
}

\begin{abstract}
In recent years, tracking human motion using IMUs from everyday devices such as smartphones and smartwatches has gained increasing popularity.
However, due to the sparsity of sensor measurements and the lack of datasets capturing human motion over uneven terrain, 
existing methods often struggle with pose estimation accuracy and are typically limited to recovering movements on flat terrain only.
To this end, we present BaroPoser, the first method that combines IMU and barometric data recorded by a smartphone and a smartwatch to estimate human pose and global translation in real time. 
By leveraging barometric readings, we estimate sensor height changes, which provide valuable cues for both improving the accuracy of human pose estimation and predicting global translation on non-flat terrain.
Furthermore, we propose a local thigh coordinate frame to disentangle local and global motion input for better pose representation learning.
We evaluate our method on both public benchmark datasets and real-world recordings. 
Quantitative and qualitative results demonstrate that our approach outperforms the state-of-the-art (SOTA) methods that use IMUs only with the same hardware configuration.

\end{abstract}


\begin{teaserfigure}
  \includegraphics[width=\textwidth]{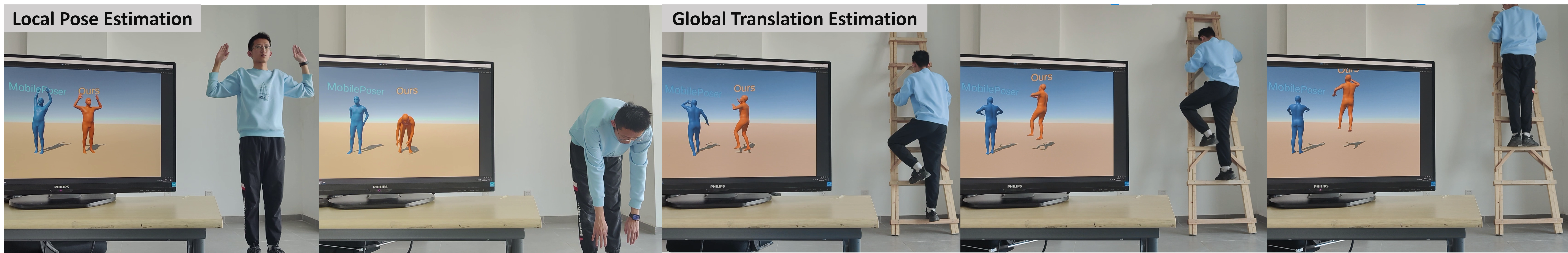}
  \caption{Live comparison between prior work~\cite{xu2024mobileposer} (blue) and BaroPoser (orange) on pose and translation estimation. BaroPoser leverages barometric information to improve pose accuracy and enables reconstruction of human motions involving height variations.}
  \label{fig:teaser}
\end{teaserfigure}

\maketitle

\section{INTRODUCTION}
Human motion capture (MoCap) is a long-standing and challenging problem in computer graphics and vision.
It aims to reconstruct 3D human body movements, with many applications in movie production, gaming, and AR/VR.

Although vision-based methods~\cite{peng2021neural, wenbinlin_vision, wang2024intrinsicavatar, xiu2024puzzleavatar} have made significant progress in this field, 
they always require camera visibility and are sensitive to occlusions and lighting conditions.
Some works have focused on tracking human motion via accelerations and rotations recorded by body-worn Inertial Measurement Units (IMUs), which overcome the aforementioned limitations.
Commercial solutions in this category require 17 or more IMUs, which can be intrusive and time-consuming for usage.
Recent studies~\cite{huang2018deep, yi2021transpose, yi2022physical, jiang2022transformer, van2023diffusion, yi2024physical, armani2024ultra} have reduced the number of IMUs to six or fewer, striking a balance between accuracy and practicality.
However, these methods still require specialized IMU sensors, limiting their application in everyday life.
To tackle this problem, some studies~\cite{mollyn2023imuposer, xu2024mobileposer} leverage the IMUs already available in everyday devices (such as phones, watches, wristbands, and earbuds) for human motion capture. 
These methods define a set of typical device placement locations like the head, wrists, or pockets, and use up to three of them to estimate human pose and global translation through neural networks. 
However, the accuracy of their motion estimation remains limited, 
as the problem is inherently under-constrained due to sparse and noisy IMU measurements available in everyday settings. 
This makes it difficult to accurately recover either local body poses or global translations.

\begin{figure*}[ht]
    \centering
    \includegraphics[width=1.0\linewidth]{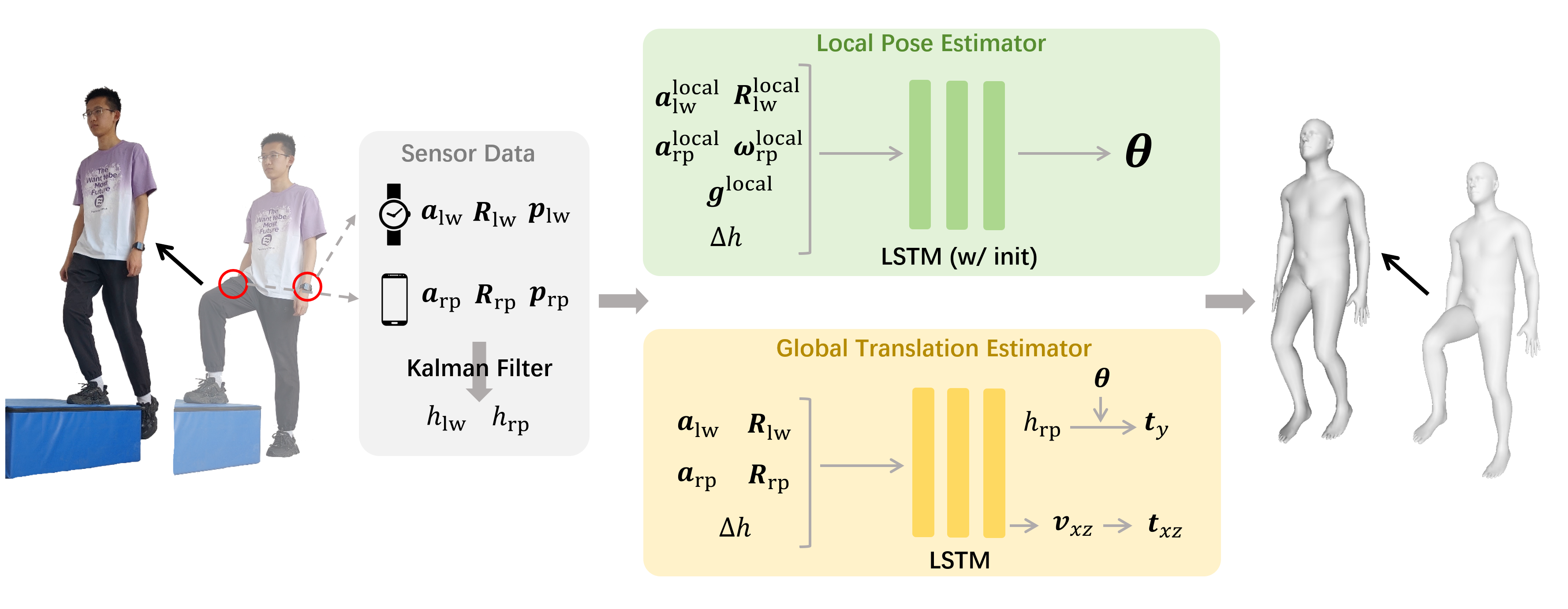}
    \caption{Overview of our method. BaroPoser takes IMU and barometric data from a smartwatch (left wrist) and a smartphone (right pocket) to estimate real-time human motion. We incorporate two core modules: (1) Local Pose Estimator, which regresses joint rotations in a local coordinate frame rooted at the thigh and incorporates relative sensor heights to reduce ambiguity; and (2) Global Translation Estimator, which predicts horizontal velocity with a neural network and vertical translation from pose-corrected thigh height. Together, these components enable accurate MoCap using only sparse sensor data from everyday devices. 
    }
    \label{fig:pipeline}
\end{figure*}

In this paper, we present BaroPoser, the first approach that fuses IMU and barometric data from one smartwatch (worn on one wrist) and one smartphone (placed in the thigh pocket of the opposite side) to estimate full-body motions in real time.
In addition to IMU data, which has been widely used for MoCap, we propose to incorporate barometric readings from built-in sensors in everyday devices such as smartphones and smartwatches. 
These readings provide information about absolute altitude, offering an additional feature to improve the accuracy of both pose and global translation estimation.
Such vertical awareness is particularly valuable in applications such as AR/VR and fitness tracking, where altitude-sensitive actions like stair climbing, squats, and jumps are common. 
Moreover, to better exploit human motion priors in this setting with only two sensors, we introduce a thigh coordinate system to represent local body poses, which helps to decouple the local and global motions.
Specifically, we define a local coordinate frame for the sensor on the thigh and treat it as the root coordinate frame of the human local pose. 
Then, both the input and output of the pose estimation network are represented in this root coordinate frame.
In this manner, the global and local motion information recorded by the sensors is disentangled naturally as the thigh sensor records the global motion information while the wrist sensor records the local one.
This allows the deep learning model to focus on learning the local motion prior without any interference from the global motions.
Otherwise, we cannot extract pure local motion signals from the sensor input to regress local poses with the help of the learned local pose prior.
We evaluate our method on multiple benchmark datasets with real IMU data and show consistent improvements over prior approaches. 
Additionally, we collect a real-world dataset and conduct quantitative and qualitative evaluations to further validate our method in practical deployment scenarios.

In summary, our main contributions include:
\begin{itemize}
    \item We propose BaroPoser, the first method that fuses IMU and barometric data from everyday devices to estimate full-body motion in real time.
    \item We introduce barometric height as an additional feature to enhance both pose and translation estimation under the two-sensor configuration.
    \item We establish a local coordinate system rooted at the thigh to remove orientation dependency and improve motion representation learning.
\end{itemize}

\section{RELATED WORKS}

\subsection{Inertial MoCap with Specialized Sensors}
Inertial Measurement Units (IMUs) are popular for human motion capture due to their small size, low power consumption, and portability.
Recently, commercial inertial motion capture methods use 17 IMUs to accurately capture human motion~\cite{Xsens2024, noitom2021perception}. 
However, this approach is costly, requiring cumbersome equipment and skilled technicians to operate.

Recent studies have reduced the number of sensors required to six or fewer.
Sparse Inertial Poser (SIP)~\cite{von2017sparse} proposed to reconstruct human pose and translation from six IMUs in an offline manner.
Deep Inertial Poser (DIP)~\cite{huang2018deep} proposed a deep learning-based method that uses six IMUs to capture human pose in real time.
Building upon this, TransPose~\cite{yi2021transpose} estimates both human pose and global translation by adopting a multi-stage pose estimation and fusion-based global translation estimation.
Subsequent work~\cite{yi2022physical, jiang2022transformer, zhang2024dynamic, yi2024physical, armani2024ultra, van2023diffusion} has further improved the accuracy and robustness of sparse inertial motion capture. 
PIP~\cite{yi2022physical} introduces a physics-aware motion optimizer, which reduces motion jitter and improves physical plausibility.
Built on Transformer~\cite{vaswani2017attention}, TIP~\cite{jiang2022transformer} additionally estimates Stationary Body Points (SBPs), which enable the reconstruction of terrian consistent with the human motion.
DynaIP~\cite{zhang2024dynamic} leverages pseudo velocities and a part-based design to improve pose estimation with enhanced robustness and consistency.
PNP~\cite{yi2024physical} uses an autoregressive model to learn fictitious forces for improved physical consistency in the non-inertial root frame.
UIP~\cite{armani2024ultra} incorporates UWB-based distance measurements into inertial pose estimation, enabling robust and scalable motion capture through distance-constrained learning.
Moreover, DiffusionPoser~\cite{van2023diffusion} employs a diffusion-based generative model with autoregressive inference, enabling motion capture with arbitrary numbers and placements of IMUs.
Global Inertial Poser~\cite{yi2025improving} improves the accuracy of both local pose and global orientation by integrating gravity priors and enables the estimation of the unconstrained motion in 3D space through 3D contact detection based on physics.

However, although these methods significantly improve the accuracy and robustness of inertial motion capture with fewer sensors, they still rely on specialized IMU devices that are costly, cumbersome, and not typically accessible to everyday users.
Moreover, most of them require six IMUs placed on specific body locations, which imposes strict hardware constraints and limits usability in casual or mobile scenarios.
As a result, the application of these techniques is largely restricted to controlled environments such as labs or studios, limiting their potential for broader real-world deployment.

\subsection{Inertial MoCap with Everyday Devices}
Earlier research~\cite{kwapisz2011activity, zeng2014convolutional, hammerla2016deep} has explored leveraging IMU data from consumer devices such as smartphones and smartwatches to identify coarse-grained activities like walking, sitting, or stair climbing. 
Some of these works~\cite{karande2024raising, graumuller2025posture} incorporate barometric sensing to estimate elevation change or detect activities involving vertical transitions. 
However, these approaches are typically limited to classifying discrete activity labels, without estimating fine-grained body pose or continuous global translation. 

Building on HAR research, recent studies have extended the use of IMU data from everyday devices to full-body motion capture. 
IMUPoser~\cite{mollyn2023imuposer} achieves real-time human pose estimation using any combination of a smartphone, a watch, and earbuds for the first time. 
Specifically, it defines five possible on-body device locations and trains a bidirectional LSTM to directly regress the human pose from any valid configurations.
Lee et al.~\cite{lee2024mocap} propose to estimate human motion from smartwatches and a head-mounted camera.
SmartPoser~\cite{devrio2023smartposer} focuses on estimating upper-limb poses using a single smartphone placed on the upper arm. 
The most relevant work to ours is MobilePoser~\cite{xu2024mobileposer}, which builds upon the multi-stage strategy of TransPose~\cite{yi2021transpose} to simultaneously estimate human pose and global translation using everyday consumer devices.

These works extend inertial motion capture to consumer devices, making the technology more practical and accessible for everyday use.
However, due to the sparsity of sensors in daily use and the lack of uneven-terrain motion data, 
the accuracy of motion capture remains limited, and existing methods are typically constrained to recovering movements on flat terrain.
To address this challenge, we leverage the observation that everyday devices such as smartphones and smartwatches not only provide IMU data but also naturally include barometric pressure readings, and propose the first method that fuses this information to enhance inertial motion capture in everyday settings.
\section{METHOD}
Our task is to estimate human pose and global translation in real time using IMU and barometer data from a smartwatch and a smartphone, which are attached to the left wrist and right thigh pocket, respectively.
We divide this task into two subtasks: height-aware local pose estimation (Section~\ref{sec: local pose estimation}) and hybrid global translation estimation (Section~\ref{sec: global transaltion estimation}).
Figure~\ref{fig:pipeline} shows an overview of our pipeline, which we describe in detail in the following sections.
\subsection{System Input}
Our system takes as input the accelerations and rotations from the IMUs, along with barometric pressure readings that indicate the sensor’s height. However, barometric measurements are highly susceptible to environmental factors such as temperature and humidity. To mitigate the noise, we employ a Kalman filter~\cite{kalman1960new} to fuse the barometric pressure data with the IMU readings, yielding a more stable and accurate height estimation.
After that, we align the inertial measurements and filtered heights to the same reference frame, resulting in a data vector $\bm{x} \in \mathbb{R}^{26}$:
\begin{equation}
\label{equation: x}
\bm{x} = [\bm{a}_{\text{lw}}, \bm{a}_{\text{rp}}, \bm{R}_{\text{lw}}, \bm{R}_{\text{rp}}, h_{\text{lw}}, h_{\text{rp}}]
\end{equation}
where $\bm{a}\in\mathbb{R}^3$ is the acceleration, $\bm{R}\in\mathbb{R}^{3\times 3}$ is the rotation matrix and $h\in\mathbb{R}^1$ is the filtered height.
The subscripts lw and rp refer to the left wrist and the right thigh pocket, respectively.

\subsection{Height-aware Local Pose Estimation}
\label{sec: local pose estimation}
In this section, we present our method for estimating human pose, i.e. joint rotations, from sparse IMUs and barometric measurements.
Previous works \cite{mollyn2023imuposer, xu2024mobileposer} take IMU measurements from consumer devices as input and employ bidirectional LSTM \cite{birnn, lstm} networks to regress human pose, with both the input and output represented in the global coordinate frame.
However, the same motion performed in different directions corresponds to the same local pose, making global-coordinate modeling prone to entangling global orientation with pose.
Furthermore, inferring full-body joint rotations from extremely sparse inertial signals remains highly under-constrained and ambiguous.

To address these issues, we establish a local coordinate system by treating the right pocket sensor as the root, and transform both the IMU measurements and the predicted joint rotations into this local frame.
In addition, we leverage the built-in barometers in everyday devices to provide the relative height between the two sensors as an extra input, which helps reduce the ambiguity in pose estimation. 
Specifically, we adopt a data-driven approach and the input to the network is a concatenated vector \( [ \bm{a}_{\text{lw}}^{\text{local}},\ \bm{a}_{\text{rp}}^{\text{local}},\ \bm{R}_{\text{lw}}^{\text{local}},\ \bm{\omega}^{\text{local}}_{\text{rp}},\ \bm{g}^{\text{local}},\ \Delta h ]  \in \mathbb{R}^{22}\), which can be derived from Eq.~\eqref{equation: x} as follows:
\begin{align}
\bm{a}_{\text{lw}}^{\text{local}} &= \bm{R}^T_{\text{rp}}\bm{a}_{\text{lw}} \\
\bm{a}_{\text{rp}}^{\text{local}} &= \bm{R}^T_{\text{rp}}\bm{a}_{\text{rp}} \\
\bm{R}_{\text{lw}}^{\text{local}} &= \bm{R}^T_{\text{rp}}\bm{R}_{\text{lw}} \\
\bm{\omega}_{\text{rp}}^{\text{local}}(t) &= \frac{1}{\Delta t}\text{Log}(\bm{R}_{\text{rp}}(t-1)^T\bm{R}_{\text{rp}}(t)) \\
\bm{g}^{\text{local}} &=\bm{R}_{\text{rp}}^T\bm{g} \\
\Delta h &= h_{\text{lw}} - h_{\text{rp}}
\end{align}
Here, $\bm{a}^{\text{local}}\in\mathbb{R}^3$, $\bm{\omega}^{\text{local}}\in\mathbb{R}^3$, and $\bm{R}^{\text{local}}\in\mathbb{R}^{3\times 3}$ represent the acceleration, angular velocity, and rotation in the root coordinate frame, respectively;
$\bm{g}^{\text{local}} \in \mathbb{R}^3$ and $\bm{g} \in \mathbb{R}^3$ is the unit gravity direction expressed in the root and world frame, respectively, which can be used to distinguish between poses such as standing and lying;
$\Delta h$ represents the height difference between the two sensors, $\Delta t=1/30$ is the time interval, and $\mathrm{Log}(\cdot) : SO(3) \to \mathbb{R}^3$ maps the rotation matrix to the vector space.
The output of the network is the SMPL~\cite{SMPL:2015} joint rotations $\theta \in \mathbb{R}^{24\times 6}$ in the 6D~\cite{zhou2019continuity} representation, also expressed in the root coordinate frame.
The loss function used to train the model is defined as follows:
\begin{equation}
\label{eq: pose loss}
\mathcal{L}_{\theta} = \left\| \theta - \theta^{\text{GT}} \right\|^2_2
\end{equation}
where superscript GT denotes the ground truth.

By transforming global measurements into the local coordinate frame, the network learns orientation-free local poses that remain consistent across different subject global orientations.
Moreover, we leverage the built-in sensors in everyday devices to incorporate a height-related feature, which augments the sparse inertial signals and improves the accuracy of pose estimation.

\subsection{Hybrid Global Translation Estimation}
\label{sec: global transaltion estimation}
In this section, we describe our method for estimating global translation.
Prior work~\cite{xu2024mobileposer} combines two sources of velocity: 
one is computed using the estimated pose and the probability of foot contact, while the other is predicted by a neural network based on concatenated IMU signals and estimated joint positions.
However, this approach has two main limitations: 
It assumes the subject is always moving on flat ground, due to the lack of datasets containing motions with vertical variation, 
and it heavily relies on accurate lower-body pose estimation, which is difficult to achieve in everyday settings where only a single IMU is available on the thigh.

To address this issue, we leverage barometric information to decompose the global translation into horizontal and vertical components.
The horizontal velocity is directly estimated by the network using only sensor data, 
while the vertical translation is derived from the filtered height at the thigh combined with the estimated pose.

Specifically, we estimate horizontal velocity using a data-driven approach based on a unidirectional LSTM. 
The input to the network is a concatenated vector \( [ \bm{a}_{\text{lw}},\ \bm{a}_{\text{rp}},\ \bm{R}_{\text{lw}}, \bm{R}_{\text{rp}}, \Delta h ]  \in \mathbb{R}^{25}\), 
where $\bm{a}\in\mathbb{R}^3$, $\bm{R}\in\mathbb{R}^{3\times 3}$, $\Delta h\in\mathbb{R}^{1}$ represent the acceleration, rotation matrix, and height difference in the global coordinate frame, respectively.
The output is the horizontal velocity $\bm{v}_{xz} \in \mathbb{R}^2$.
Inspired by \cite{yan2019ronin, zheng2024neurit}, we use a global accumulated velocity loss, which integrate predicted velocities over the entire sequence and compare them with the overall translation:
\begin{equation}
\label{eq: velocity loss}
\mathcal{L}_{v_{xz}} = \left\| \sum_{i=1}^{N}v_i - \sum_{i=1}^{N}v_i^{\text{GT}} \right\|^2_2
\end{equation}
where $N=150$ is the sequence length during training, $v$ is the horizontal velocity, and $v^{\text{GT}}$ denotes the ground truth.

For vertical translation, we approximate it using the filtered height of the thigh, which is physically close to the root. 
However, motions such as leg lifting can cause significant changes in thigh height without affecting the root. 
To correct this, we subtract the local thigh variation computed by the estimated pose from the barometer-derived global height change:
\begin{equation}
\label{eq: trans_y}
t_{y,t} = (h_{t}^{\text{glb}}-h_{0}^{\text{glb}})-(h_{t}^{\text{local}}-h_{0}^{\text{local}})
\end{equation}
where $t_y$ denotes the vertical translation of root,
the subscripts $\text{glb}, \text{local}$ refer to the thigh height obtained from the filtered barometric readings and the local pose, the subscripts $t, 0$ indicate the current and inital time steps.
By incorporating barometric information, our method is able to produce plausible non-flat terrain motions (e.g., stair climbing) without requiring additional training.
\section{IMPLEMENTATION DETAILS}
\begin{table*}[ht]
\centering
\caption{Quantitative comparison of pose estimation on public datasets.}
\resizebox{\textwidth}{!}{
\begin{tabular}{@{}lcccccccccccc@{}}
\toprule
\multirow{1}{*}{Dataset}  & \multicolumn{4}{c}{
DIP-IMU} & \multicolumn{4}{c}{IMUPoser} & \multicolumn{4}{c}{TotalCapture} \\ 
\cmidrule(lr){2-5} \cmidrule(lr){6-9} \cmidrule(lr){10-13}
\multirow{1}{*}{Method} & SIP Err & Ang Err & Pos Err & Mesh Err & SIP Err & Ang Err & Pos Err & Mesh Err & SIP Err & Ang Err & Pos Err & Mesh Err \\ 
\midrule
IMUPoser  & 27.21 & 23.10 & 9.99 & 12.34 & 18.20 & 20.39 & 8.92 & 11.56 & 17.38 & 18.79 & 8.92 & 10.78 \\ 
MobilePoser  & 26.94 & 22.39 & 9.33 & 11.33 & 17.52 & 19.51 & 8.57 & 11.02 & 17.20 & 18.59 & 8.78 & 10.52 \\ 
Ours & \textbf{24.39} & \textbf{21.25} & \textbf{8.57} & \textbf{10.36} & \textbf{13.15} & \textbf{16.20} & \textbf{6.82} & \textbf{8.95} & \textbf{14.52} & \textbf{16.15} & \textbf{7.36} & \textbf{8.94} \\ 
\bottomrule
\end{tabular}
}
\label{tab:pose quantitative results}
\end{table*}
\begin{table}[h]
\centering
\caption{Quantitative comparison of pose estimation on real-world dataset.}
\label{tab:real_pose}
\begin{tabular}{lcccc}
\toprule
Method & SIP Err  & Ang Err  & Pos Err  & Mesh Err  \\
\midrule
IMUPoser & 28.49 & 26.50 & 12.24 & 14.85 \\
MobilePoser & 28.01 & 25.74 & 12.00 & 14.46 \\
Ours & \textbf{24.76} & \textbf{23.53} & \textbf{10.73} & \textbf{12.97} \\
\bottomrule
\end{tabular}
\end{table}
\subsection{Data Processing and Collection} 
\begin{figure}[ht]
    \centering
    \includegraphics[width=\linewidth]{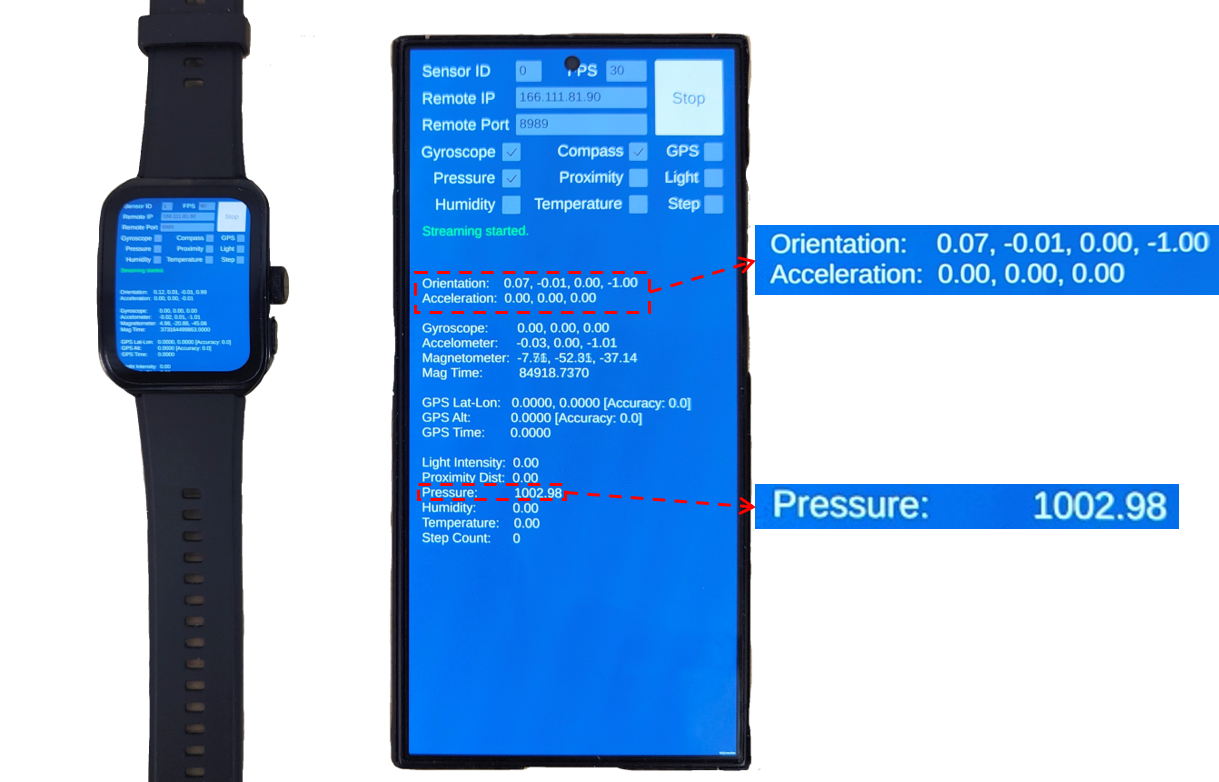}
    \caption{Hardware setup for collecting real IMU and barometric data. We developed an Android-based app for collecting sensor data from both the smartwatch and the smartphone. Users can input an IP address to send the collected IMU and barometric data to a terminal.}
    \label{fig:sensors}
\end{figure}

Following prior work~\cite{mollyn2023imuposer, xu2024mobileposer}, we train our model on the AMASS~\cite{mahmood2019amass} dataset and evaluate it on DIP-IMU, IMUPoser, and TotalCapture datasets. 
During training, we use synthetic IMU accelerations and rotations, along with the heights of two locations (the left wrist and right pocket) relative to the initial ground, which is defined based on the lowest point of the feet in the first frame.
To simulate real-world sensor variability, we inject Gaussian noise into the synthesized altitude data to improve the model’s robustness to barometric fluctuations (Gaussian noise with std = 0.05m).
At test time, real IMU accelerations and rotations are used, while the heights are synthetically generated as in training.
We also enable real-time IMU and barometer data collection using an OPPO Watch 4 Pro and a Galaxy S24 Ultra, as shown in Figure~\ref{fig:sensors}, for live demos and evaluations on real-world data.
As the consumer devices used in our system capture data at 30 FPS, we resample both the training and testing data to match this frame rate.
\subsection{Sensor Calibration}
\begin{figure}[ht]
    \centering
    \includegraphics[width=0.8\linewidth]{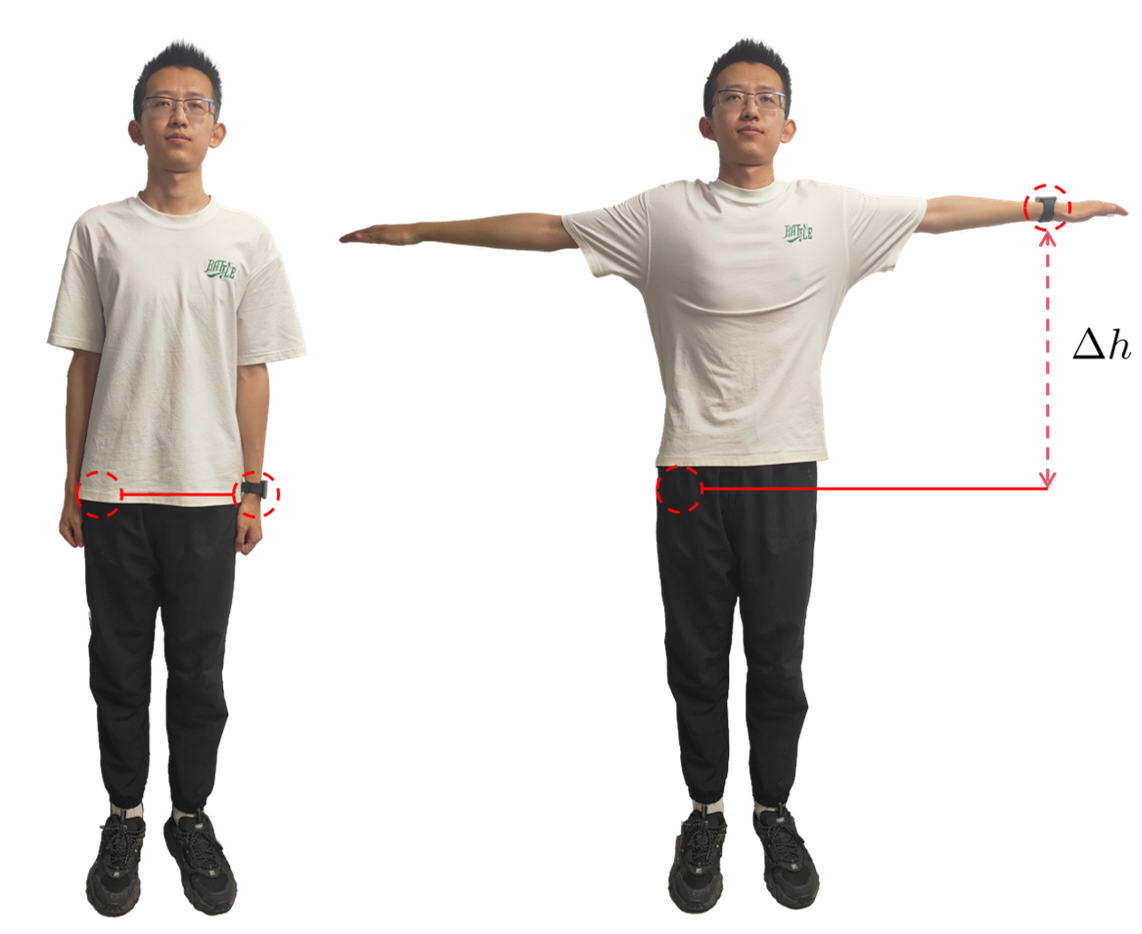}
    \caption{Sensor calibration procedure. Left: Both devices (watch and phone) are placed at the same height to determine barometric bias. Right: A T-pose is performed to establish a known vertical offset $\Delta h$ (0.66m) , used to compute the pressure-to-height mapping scale.}
    \label{fig:calibration}
\end{figure}

Due to hardware differences and environmental sensitivity, barometric sensors may exhibit biases and inconsistent pressure-to-height scales. 
Therefore, we calibrate the barometric sensors on both devices to correct for bias and scale.
Specifically, we first place the smartphone and smartwatch at approximately the same height to compute the relative bias between their barometric sensors after wearing the devices. 
We then perform a T-pose, where the relative height between the two devices is known by assuming a mean body shape, which enables us to determine the pressure-to-height scaling factor in the current environment. 
The full calibration process is illustrated in Figure~\ref{fig:calibration}.
\begin{figure}[ht]
    \centering
    \includegraphics[width=0.8\linewidth]{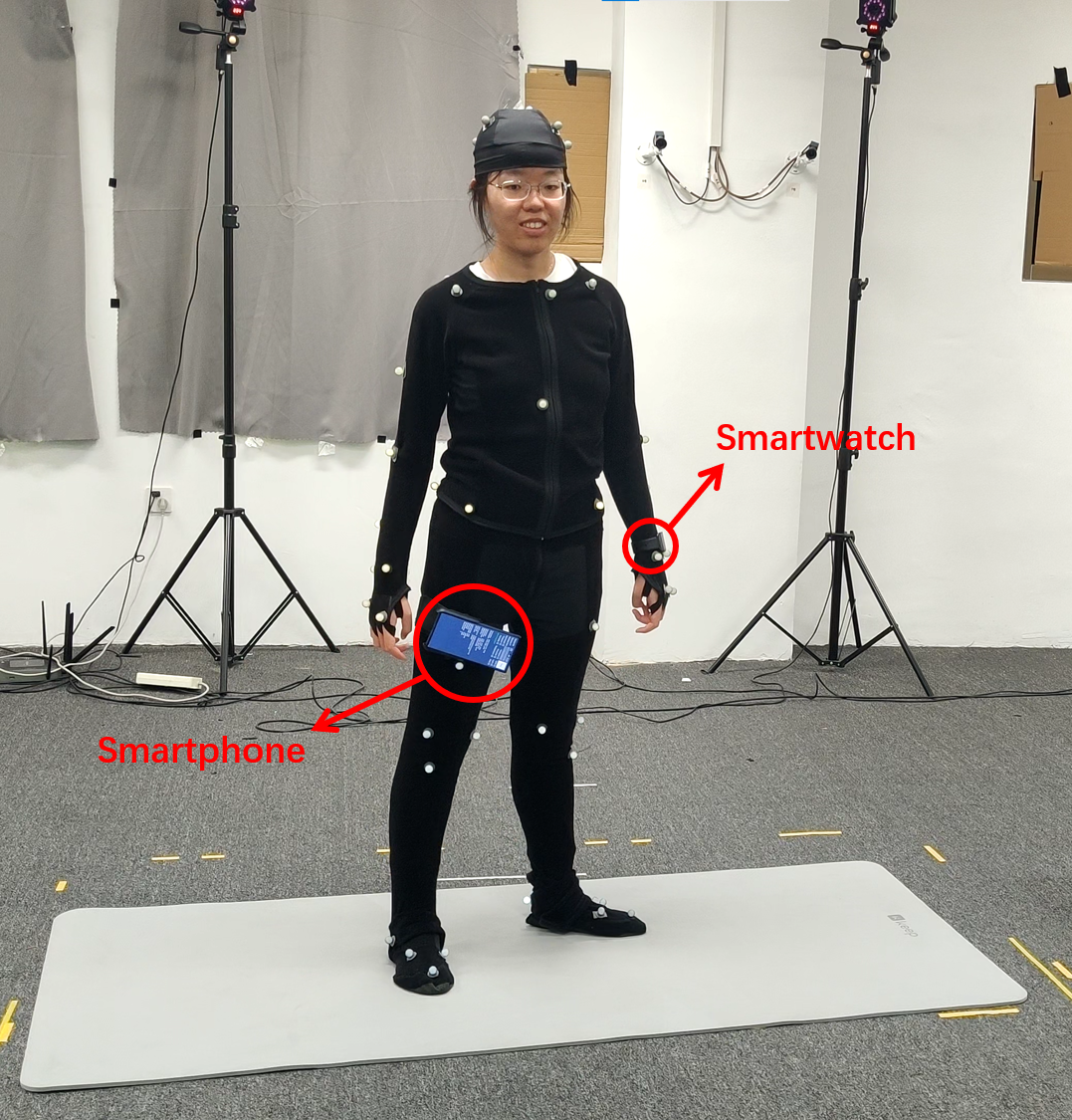}
    \caption{Illustration of our data collection setup. Subjects wear reflective markers for motion capture, while a smartphone is placed in the right thigh and a smartwatch is worn on the left wrist. The motion data is recorded using a 20-camera infrared motion capture system.}
    \label{fig:data_collection}
\end{figure}
\subsection{Network Structure}
For the pose regression network, we use a 3-layer MLP with the hidden size of 512 to encode the first-frame pose into the initial LSTM hidden state, followed by a 2-layer unidirectional LSTM with the hidden size of 512.
Inspired by \cite{yi2022physical}, we use a 3-layer MLP to initialize the hidden state of the LSTM in a learnable manner, which helps the model adapt to different initial poses.
For horizontal velocity regression, we directly use a 2-layer unidirectional LSTM with a hidden size of 512.
All networks are trained on a single NVIDIA RTX 4090 GPU for 100 epochs with a batch size of 256, using the Adam~\cite{kingma2014adam} optimizer with a learning rate of 3e-4. The training takes about 6 hours.
\section{EXPERIMENTS}
In this section, we first introduces the datasets used in our experiments (Section~\ref{sec: datasets}), 
followed by both quantitative and qualitative results for pose (Section~\ref{sec: exp pose}) and translation estimation (Section~\ref{sec: exp global translation}). 
We further perform ablation studies to validate the effectiveness of our key technical components (Section~\ref{sec:ablation}).

\subsection{Datasets}
\label{sec: datasets}
We conduct both quantitative and qualitative experiments on three public datasets containing real IMU data:
\begin{itemize}
    \item \textit{DIP-IMU}~\cite{huang2018deep} contains data from 10 participants, captured with commercial-grade Xsens\cite{Xsens2024} IMUs at 60 fps, but it does not include global translation data.
    \item \textit{TotalCapture}~\cite{totalcapture} provides real IMU data with corresponding ground-truth pose and translation, captured using Xsens IMUs at 60 fps.
    \item \textit{IMUPoser}~\cite{mollyn2023imuposer} includes real IMU data paired with ground-truth pose and translation data captured using consumer-grade devices: an iPhone 11 Pro, Apple Watch Series 6, and AirPods, at 25 fps.
\end{itemize}
As these datasets do not include barometric data, we synthesize height information at the left wrist and right pocket.

In addition to public datasets, we collected a real-world dataset using a smartphone and a smartwatch, which record synchronized IMU and barometric measurements.
To obtain ground-truth pose and translation, we use a motion capture system with 20 infrared cameras and placed reflective markers on the subject’s body.  
The smartphone and smartwatch are securely attached to the right thigh and left wrist, respectively.
An illustration of our data collection setup is shown in Figure~\ref{fig:data_collection}.
The dataset includes approximately 30 minutes of motion data from 5 subjects performing 10 actions, covering a range of height-changing activities such as stair stepping and ladder climbing.
We also use this dataset to evaluate both pose estimation and global translation accuracy quantitatively and qualitatively.

\subsection{Pose Estimation Evaluation}
\label{sec: exp pose}
\subsubsection{Evaluation Metrics}
We use the following metrics to quantitatively evaluate pose estimation: 
\begin{itemize}
    \item \textit{SIP error} measures the mean global rotation error of upper arms and upper legs in degrees.
    \item \textit{Angular error} measures the mean global rotation error of all body joints in degrees.
    \item \textit{Positional error} measures the mean Euclidean distance error of all estimated joints in centimeters with the root joint aligned.
    \item \textit{Mesh error} measures the mean Euclidean distance between the predicted and ground-truth mesh vertices in centimeters after root alignment.
\end{itemize}
\subsubsection{Quantitative comparisons}
We conduct a quantitative comparison with IMUPoser~\cite{mollyn2023imuposer} and MobilePoser~\cite{xu2024mobileposer}, two inertial motion capture methods that also rely on everyday devices.
To ensure a fair comparison, we follow a consistent setup across all methods, using devices at the left wrist and right thigh pocket, and performing evaluation in an online manner.
The results on the three public datasets are shown in Table~\ref{tab:pose quantitative results}.
Our method achieves significant improvements in pose estimation performance across all three datasets, which can be attributed to the use of local coordinate representation and height features.
To further validate our method in practical real-world conditions, we also evaluate it on a newly collected dataset with real IMU and barometer readings. 
As shown in Table~\ref{tab:real_pose}, our full model outperforms prior methods across all metrics, confirming the effectiveness and generalizability of our approach in real deployment scenarios.
\subsubsection{Qualitative comparisons}
\begin{figure}[ht]
    \centering
    \includegraphics[width=\linewidth]{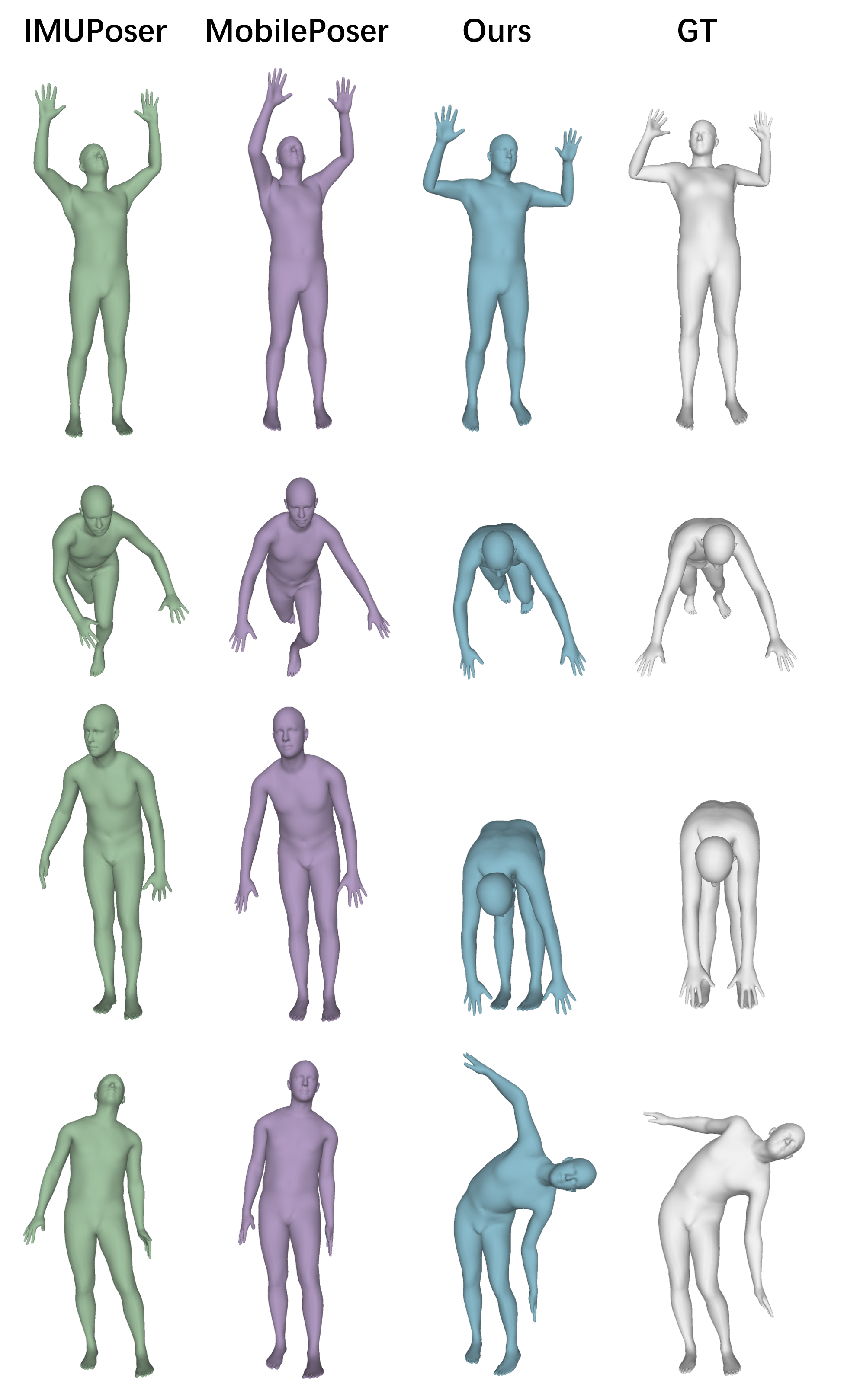}
    \caption{Qualitative comparison of pose estimation on DIP-IMU, IMUPoser, and TotalCapture datasets.}
    \label{fig:pose qualitative}
\end{figure}

Figure~\ref{fig:pose qualitative} shows a qualitative comparison between our method and prior works. 
These examples are selected from the DIP-IMU, IMUPoser, and TotalCapture datasets.
It can be observed that our method can achieves more accurate human pose estimation and performs particularly well on more challenging motions such as push-ups (the second row) and forward bends (the third row).

\begin{figure}[ht]
    \centering
    \includegraphics[width=\linewidth]{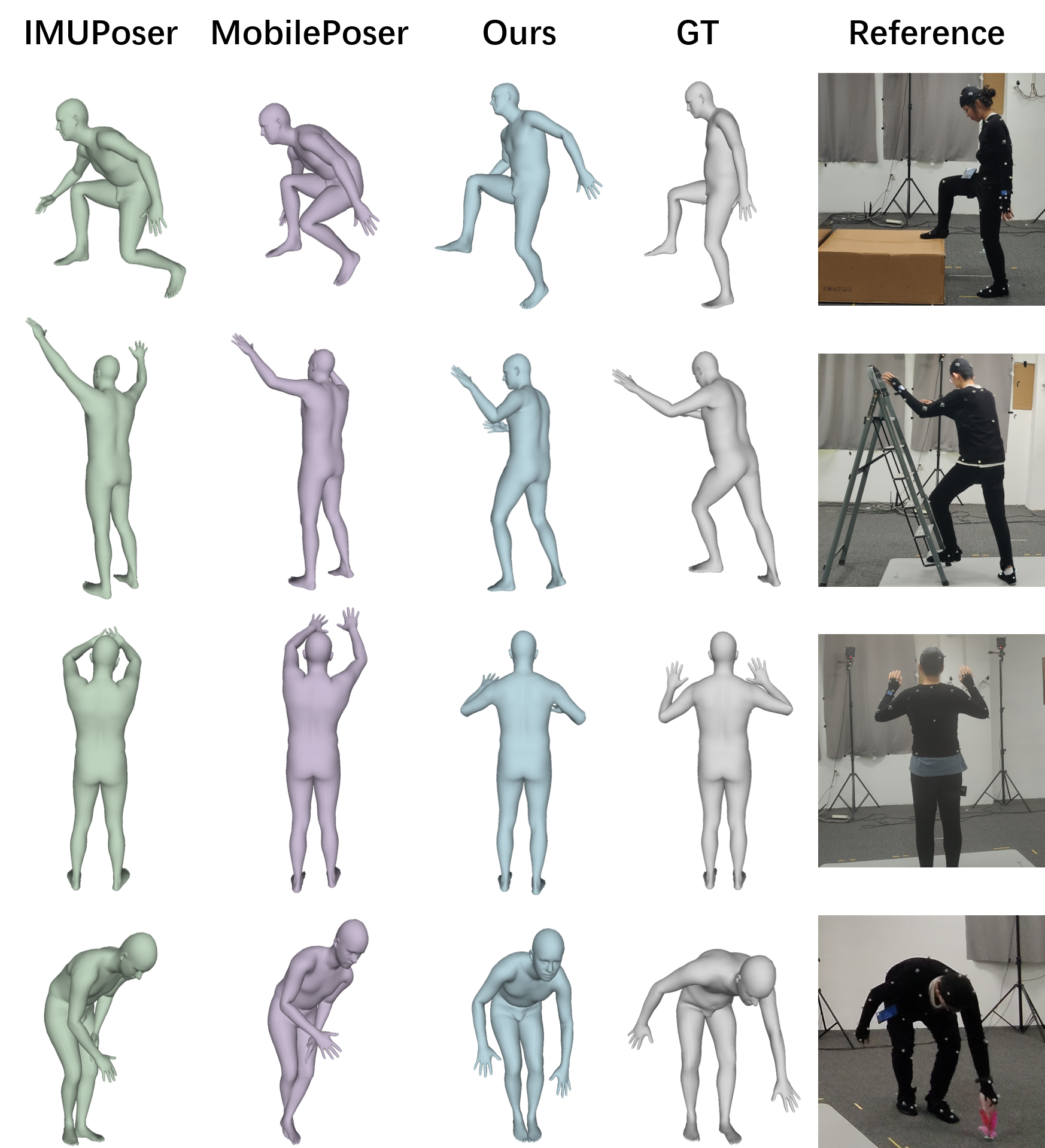}
    \caption{Qualitative comparison of pose estimation on Real-World datasets.}
    \label{fig:real pose qualitative}
\end{figure}

In addition, Figure~\ref{fig:real pose qualitative} presents qualitative results on our real-world dataset collected from everyday devices.
Our method successfully reconstructs full-body motions, including those involving vertical displacement such as stair climbing and forward bending, demonstrating its effectiveness under real-world conditions.
\begin{figure}[ht]
    \centering
    \includegraphics[width=\linewidth]{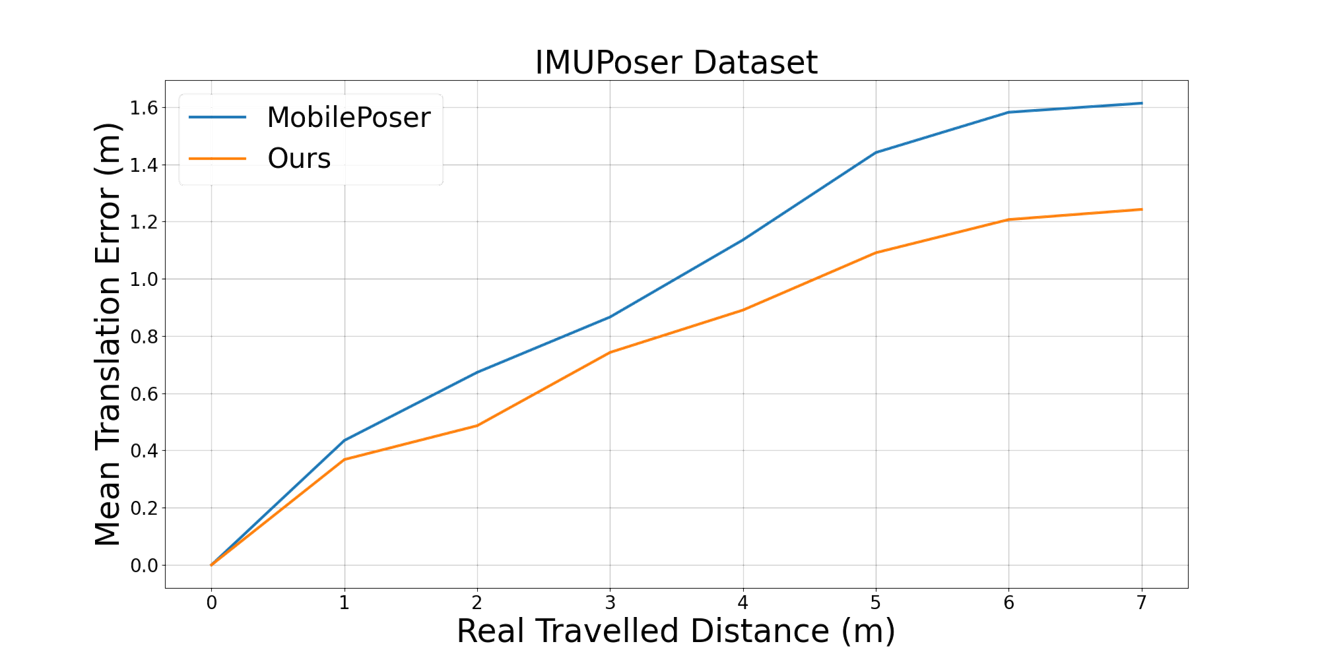}
    \includegraphics[width=\linewidth]{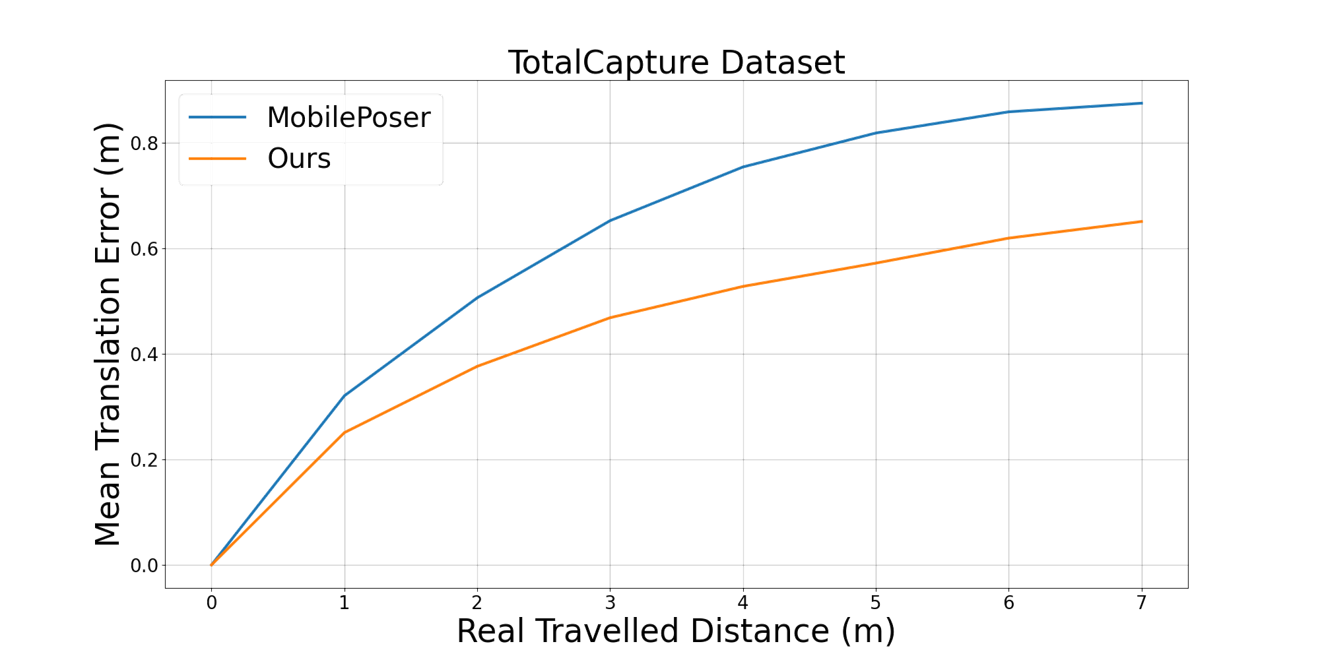}
    \includegraphics[width=\linewidth]{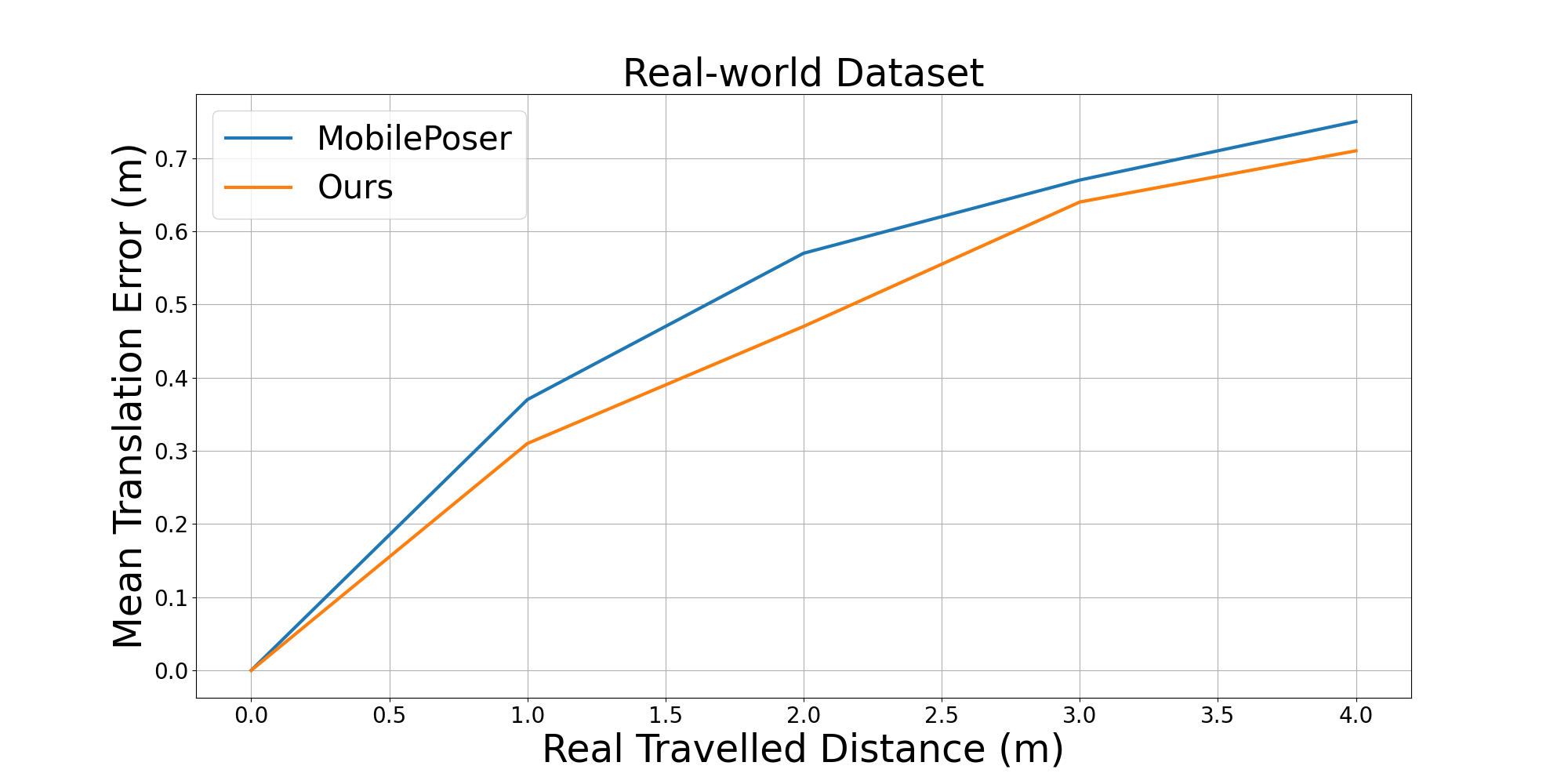}
    \caption{Quantitative comparison of translation estimation on IMUPoser, TotalCapture and real-world datasets. The x-axis represents the real distance traveled in meters, and the y-axis shows the mean translation error in meters.}
    \label{fig:tran quantitative}
\end{figure}
\subsection{Translation Estimation Evaluation}
\label{sec: exp global translation}
\subsubsection{Quantitative Comparisons}
We evaluate global translation estimation on the TotalCapture, IMUPoser and real-world datasets (as the DIP-IMU dataset does not provide ground-truth translation), and compare our method with MobilePoser~\cite{xu2024mobileposer}.
For the quantitative evaluation, we report the cumulative translation error, as shown in Figure~\ref{fig:tran quantitative}, which measures the global translation error relative to the actual distance traveled.
As illustrated, our method outperforms the previous approach on global translation estimation across both real-IMU datasets, achieving a notable improvement.
\subsubsection{Qualitative Comparisons}
\begin{figure}[ht]
    \centering
    \includegraphics[width=0.9\linewidth]{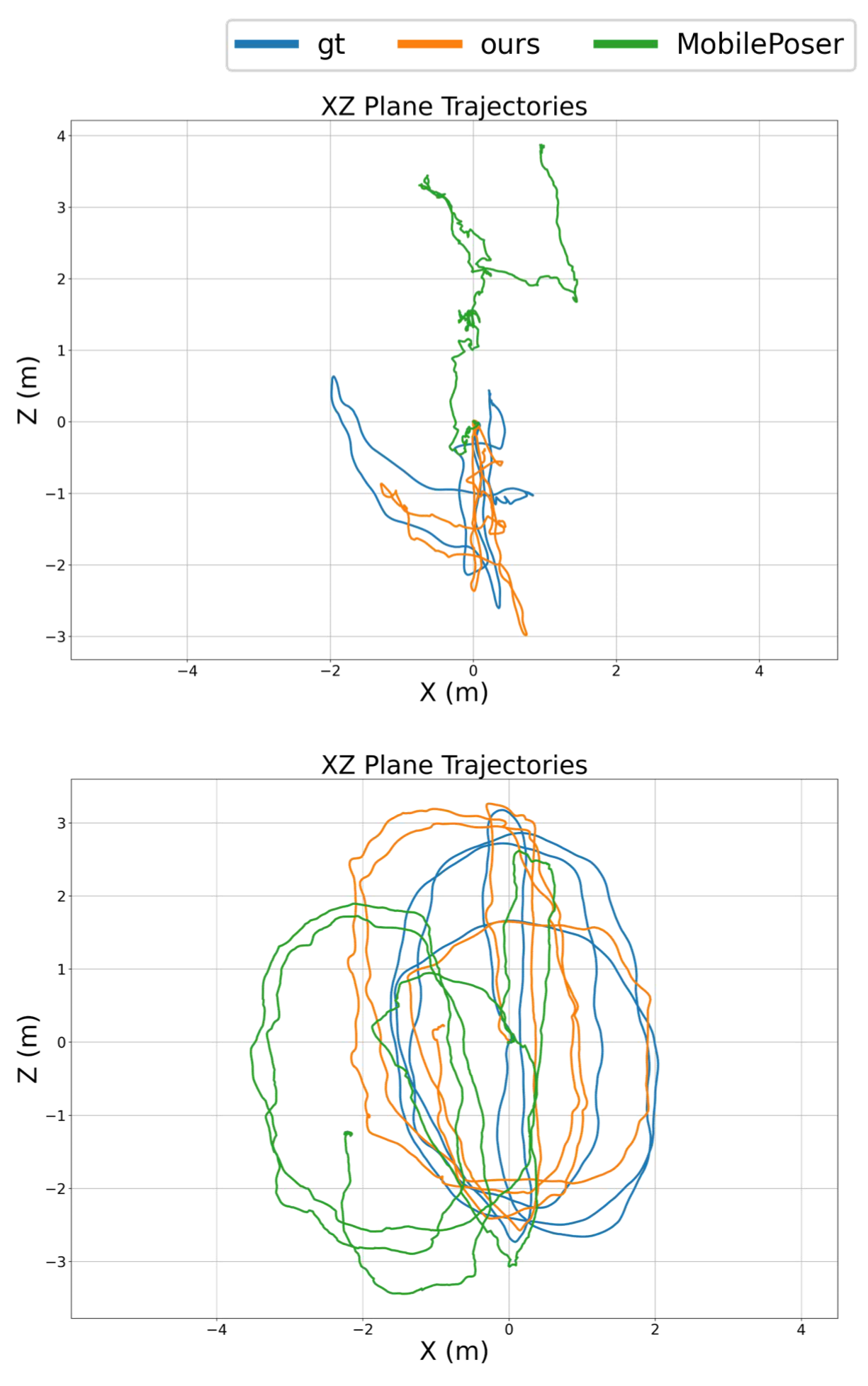}
    \caption{Qualitative comparison of translation estimation on TotalCapture dataset. Trajectories are shown on the horizontal plane and aligned at the origin.}
    \label{fig:tran qualitative}
\end{figure}

We visualize the trajectories of different methods compared to the ground truth for qualitative evaluation, as shown in Figure~\ref{fig:tran qualitative}.
The two examples are selected from the TotalCapture dataset.
Compared to previous approach, our method achieves translation results that more closely match the ground truth.

\subsection{Ablation Studies}
\label{sec:ablation}
We perform ablation studies to validate the effectiveness of the key components in our method.
For local pose estimation, we analyze the influence of adopting a thigh-rooted local coordinate frame and the use of height features.
We train our model under two modified settings: one without height input (w/o $\Delta h$), the other using global frame (w/o local) during training and inference.
As reported in Table~\ref{tab: pose ablation}, each individual modification leads to a certain improvement in performance, and when combined, our final method achieves the superior results.
The improvement introduced by adding height features is particularly evident on IMUPoser and TotalCapture datasets, which contain a wider variety of motion types.

\begin{figure}[ht]
    \centering
    \includegraphics[width=\linewidth]{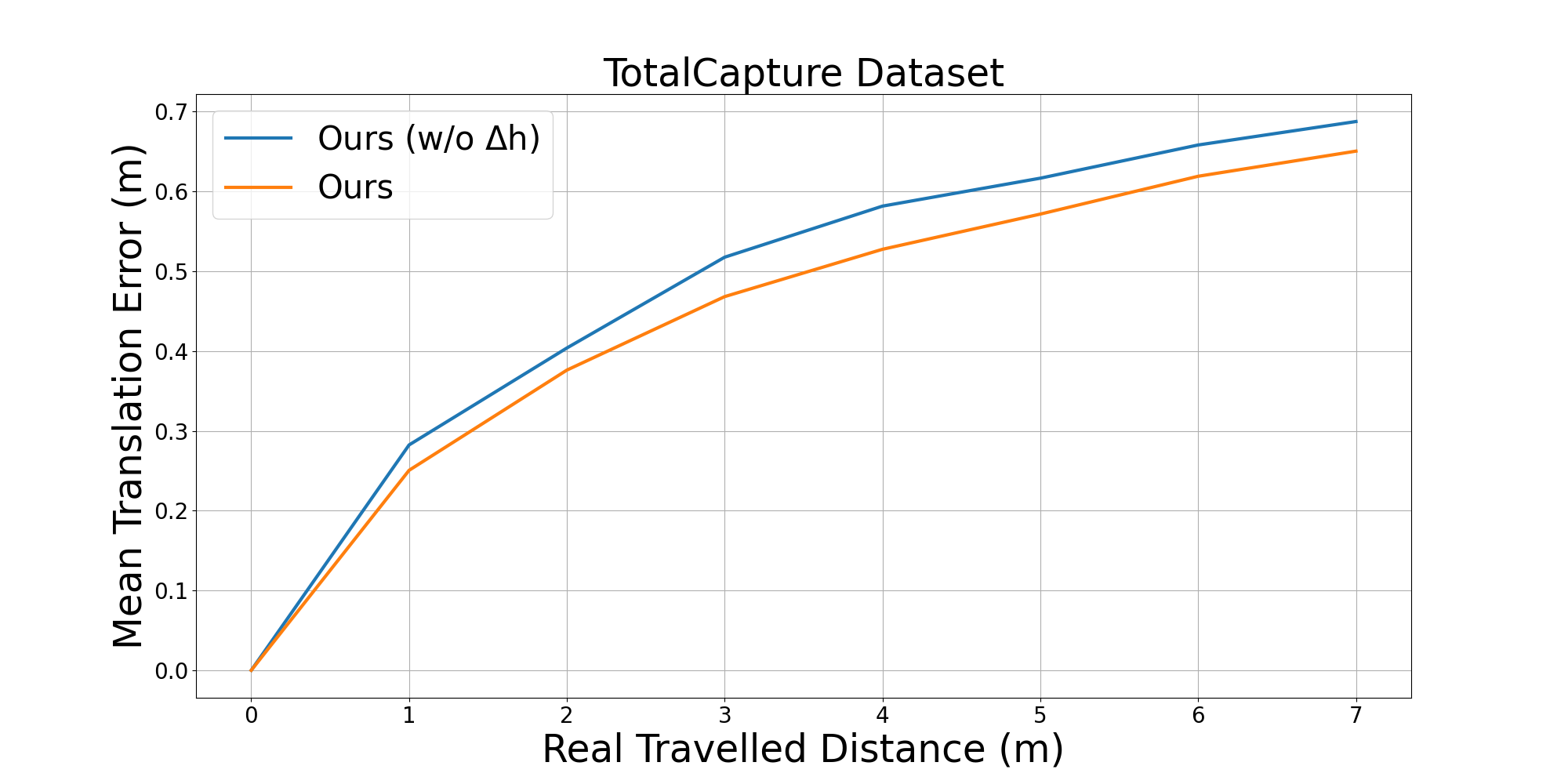}
    \caption{Ablation study on the effects of height difference for global translation estimation.}.
    \label{fig:tran ablation}
\end{figure}

For global translation estimation, we evaluate the impact of incorporating height features on the final results.
To this end, we train models with and without the height difference and quantitatively evaluate them on the TotalCapture Dataset.
As shown in Figure~\ref{fig:tran ablation}, incorporating height information also improves global translation estimation, because it provides additional cues about vertical motion and sensor dynamics.
\begin{table}[htbp]
    \centering
    \caption{Ablation study on the effects of height difference and local coordinate representation for pose estimation.}
    \label{tab: pose ablation}
    \begin{tabular}{lcccc}
        \toprule
        Method & SIP Err & Ang Err & Pos Err & Mesh Err \\
        \midrule
        \multicolumn{5}{c}{DIP-IMU} \\
        \midrule
        w/o $\Delta h$                    & 24.50 & 21.71 & 8.82 & 10.61\\
        w/o local                         & 25.33 & 22.30 & 9.12 & 11.14\\
        \textbf{BaroPoser (Ours)}         & \textbf{24.39} & \textbf{21.25} & \textbf{8.57} & \textbf{10.36}\\
        \midrule
        \multicolumn{5}{c}{IMUPoser} \\
        \midrule
        w/o $\Delta h$                    & 14.87 & 17.48 & 7.62 & 9.89\\
        w/o local                         & 15.84 & 17.76 & 7.84 & 10.08\\
        \textbf{BaroPoser (Ours)}         & \textbf{13.15} & \textbf{16.20} & \textbf{6.82} & \textbf{8.95}\\
        \midrule
        \multicolumn{5}{c}{TotalCapture} \\
        \midrule
        w/o $\Delta h$                    & 16.23 & 17.65 & 8.15 & 9.83\\
        w/o local                         & 15.68 & 17.25 & 7.99 & 9.64\\
        \textbf{BaroPoser (Ours)}         & \textbf{14.52} & \textbf{16.15} & \textbf{7.36} & \textbf{8.94}\\
        \midrule
        \multicolumn{5}{c}{Real-World Dataset} \\
        \midrule
        w/o $\Delta h$                    & 25.68 & 23.82 & 11.42 & 13.76\\
        w/o local                         & 26.08 & 25.03 & 11.56 & 13.96\\
        \textbf{BaroPoser (Ours)}         & \textbf{24.76} & \textbf{23.53} & \textbf{10.73} & \textbf{12.97}\\
        \bottomrule
    \end{tabular}
\end{table}

\section{LIMITATIONS AND FUTURE WORK}
Although BaroPoser demonstrates notable improvements in pose and translation estimation by fusing IMU and barometric readings, several limitations still remain.
First, we assume a mean body shape and fixed sensor placement on the left wrist and right thigh, which may limit its generalizability to more diverse shapes or user-defined configurations. Future work could explore enhancing robustness to flexible sensor placements and incorporating body shape information.

Another limitation is that, although we apply Kalman filtering with IMU fusion to smooth the barometric signal, barometers remain susceptible to environmental factors. 
This can lead to drift or fluctuations in the estimated vertical translation. 
One potential direction is to explore more robust sensor fusion techniques or environment-adaptive filtering to enhance the robustness of barometric measurements.

Finally, a further limitation lies in the lack of datasets that combine real IMU and barometer data with ground-truth human motion annotations, particularly for movements over uneven terrain. 
In our experiments, we perform quantitative evaluations on test sets with synthetic height information and qualitatively demonstrate real-world performance on real data. In the future, we plan to collect real-world datasets that include both barometric readings and accurate motion ground truth across diverse environments.

\section{CONCLUSION}
In this work, we present BaroPoser, the first method that fuses IMU and barometric data from a smartwatch and a smartphone to estimate full-body human pose and global translation in real time. 
To overcome the challenges posed by sparse sensor configurations, we propose a height-aware local pose estimation approach that employs a thigh-rooted local representation and incorporates the sensor height difference as an additional input to the pose regression network.
For global translation, we adopt a hybrid approach that decomposes motion into horizontal and vertical components, estimated via a neural network and pose-corrected barometric height changes, respectively.
Extensive evaluations on public and real-device datasets demonstrate that BaroPoser outperforms prior methods in both pose and translation accuracy and supports motion estimation with vertical variation. 
This work highlights the potential of integrating barometric information to expand the applicability of inertial motion capture in unconstrained real-world settings.

\section{ACKNOWLEDGMENTS}
This work was supported by the National Key R\&D Program of China (2023YFC3305600), the Zhejiang Provincial Natural Science Foundation (LDT23F02024F02), and the NSFC (No.61822111, 62021002). This work was also supported by THUIBCS, Tsinghua University, and BLBCI, Beijing Municipal Education Commission. Feng Xu is the corresponding author.

\bibliographystyle{ACM-Reference-Format}
\bibliography{sample-base}

\end{document}